\documentclass[letterpaper, 10 pt, conference]{ieeeconf}  

\IEEEoverridecommandlockouts                              

\usepackage{booktabs}
\usepackage{multirow}
\usepackage{subfigure} 
\usepackage{booktabs}
\usepackage{graphicx}
\usepackage[namelimits]{amsmath} 
\usepackage{amssymb}             
\usepackage{amsfonts}            
\usepackage{mathrsfs}            
\usepackage{algorithm2e}
\usepackage{algpseudocode}
\usepackage{url}
\usepackage{hyperref}
\hypersetup{
    colorlinks=true,
    linkcolor=blue,
    filecolor=magenta,      
    urlcolor=cyan
}

\title{\LARGE \bf
Learning Interaction-aware Motion Prediction Model for Decision-making in Autonomous Driving}

\author{Zhiyu Huang, Haochen Liu, Jingda Wu, Wenhui Huang, and Chen Lv$^{*}$,~\IEEEmembership{Senior Member, IEEE}
\thanks{Code is available at: {\href{https://github.com/MCZhi/Predictive-Decision}{https://github.com/MCZhi/Predictive-Decision}}}
\thanks{All authors are with the School of Mechanical and Aerospace Engineering, Nanyang Technological University, 639798, Singapore. (E-mails:  {\tt \{zhiyu001,  haochen002, jingda001, wenhui001\}
@e.ntu.edu.sg}, {\tt lyuchen@ntu.edu.sg})}%
\thanks{This work was supported in part by A*STAR AME Young Individual Research Grant (No. A2084c0156), and SUG-NAP Grant of Nanyang Technological University, Singapore.}
\thanks{$^{*}$Corresponding author: C. Lv}
}

\begin{document}
\maketitle
\thispagestyle{empty}
\pagestyle{empty}

\begin{abstract}
Predicting the behaviors of other road users is crucial to safe and intelligent decision-making for autonomous vehicles (AVs). However, most motion prediction models ignore the influence of the AV's actions and the planning module has to treat other agents as unalterable moving obstacles. To address this problem, this paper proposes an interaction-aware motion prediction model that is able to predict other agents' future trajectories according to the ego agent's future plan, i.e., their reactions to the ego's actions. Specifically, we employ Transformers to effectively encode the driving scene and incorporate the AV's future plan in decoding the predicted trajectories. To train the model to accurately predict the reactions of other agents, we develop an online learning framework, where the ego agent explores the environment and collects other agents' reactions to itself. We validate the decision-making and learning framework in three highly interactive simulated driving scenarios. The results reveal that our decision-making method significantly outperforms the reinforcement learning methods in terms of data efficiency and performance. We also find that using the interaction-aware model can bring better performance than the non-interaction-aware model and the exploration process helps improve the success rate in testing.
\end{abstract}

\section{INTRODUCTION}
Making safe, intelligent, and socially compatible decisions is one of the core capabilities for autonomous vehicles (AVs) targeting widespread deployment in the real world \cite{huang2021driving}. In recent years, learning-based decision-making methods have enjoyed tremendous growth and exhibited a wealth of development in autonomous driving applications, thanks to their excellent scalability and ability to handle various complex scenarios. One representative branch of learning-based methods is to directly learn a driving policy through reinforcement learning (RL) \cite{wu2022toward, huang2022efficient, wu2022prioritized} or imitation learning (IL) \cite{huang2020multi, bansal2018chauffeurnet}. However, for autonomous driving tasks, the lack of interpretability and safety guarantee of policy learning methods would hamper their applicability in practical deployment. This gives rise to the model learning approach, where a driving agent learns to predict the environment dynamics or other traffic participants’ future motions and make decisions according to the model prediction, bringing better interpretability and robustness.

Nonetheless, building the prediction model is remarkably challenging because of the complicated dependencies of agents' behaviors on the road structure and interactions among surrounding agents \cite{huang2022recoat, mo2022stochastic, liu2022strajnet}. Classical model-based methods learn a one-step transition dynamics of the environment given the agent's action \cite{sobal2022separating, henaff2019model, espinoza2022deep} and obtain a sequence of predicted environment states in an autoregressive manner. One prominent hurdle of such methods is the low prediction accuracy due to compounding errors during the rollout of the model. On the other hand, abundant research has been conducted on motion prediction models that directly forecast the agent's states several seconds into the future based on historical observations and corresponding maps. Many state-of-the-art prediction models use Transformers or graph neural networks (GNNs) \cite{ mo2022multi, huang2022multi, ngiam2021scene} to encode agent-agent and agent-map interactions have achieved outstanding prediction accuracy. Moreover, the future states of agents can be obtained via one forward pass, which could significantly accelerate the decision-making process and reduce latency. However, the main drawback of existing motion prediction models is that they ignore the influence of the AV's actions on other (environmental) agents, and the AV has to respond to the prediction results passively if using such a prediction model, which could lead to over-conservative decisions and even unsafe behaviors.

In this paper, we focus on learning an interaction-aware motion prediction model that incorporates the AV's potential future actions, so that the prediction model can respond to the AV's internal plans. From a game theory perspective, our model treats the AV as a leader and other agents as followers and predicts the responses of other agents to the AV's actions. This helps the AV reason the influence of its actions, evaluate the risks, and select the optimal one to execute. We also develop an online learning framework that enables the agent to explore and elicit the responses of other agents and such interactions are stored in a replay buffer to efficiently train the prediction model. In summary, the main contributions of this paper are listed as follows. 

\begin{enumerate}
\item We propose an interaction-aware motion prediction model to enable reactive planning and an online learning framework to train the prediction model.
\item We validate the proposed framework on several challenging simulated driving scenarios and our method shows interactive prediction results and delivers significantly better data efficiency and performance compared to baseline methods.
\item We investigate the effects of key components in our proposed motion predictor and learning framework on training and testing performance.
\end{enumerate}

\section{RELATED WORK}
\subsection{Decision-making of Autonomous Driving}
The learning-based decision-making methods for solving autonomous driving tasks can be divided into two categories. The first is \textbf{policy learning}, which focuses on training a policy (usually parameterized by neural networks) that directly outputs the driving decision given the state input. Two typical policy learning approaches are IL and RL, which have seen widespread use in autonomous driving research. IL aims to imitate the expert's decisions from their demonstrations \cite{huang2020multi, bansal2018chauffeurnet, chenIL}. However, an inevitable problem with IL is the distributional shift in online deployment, which leads to inferior testing performance. In contrast, the (model-free) RL method learns online through interactions with the environment \cite{wu2022prioritized, kiran2021deep, liu2022improved, liu2022augmenting}, which alleviates the distributional shift issue, but such trial-and-error learning can be very inefficient. Some model-based RL approaches attempt to build a transition model of dynamics and reward function and then use it to learn or improve a policy \cite{wu2022uncertainty, luo2018algorithmic, kaiser2019model}, which still fall into this category. \textbf{Model learning} is another promising line of research that learns a model to predict the environment dynamics and plan over the model \cite{hamrick2020role, moerland2020model, luo2022survey}, which improves the explainability, robustness, and safety of the system compared to policy learning. Many works follow the framework of Markov decision process (MDP) and build a one-step transition model. However, planning requires a multi-step look-ahead, and the predicted state has to be repeatedly fed back into the transition model to make long-range predictions. This could result in low prediction accuracy and divergence from the true dynamics due to accumulating errors, consequently in compromised planning performance. To tackle this problem, we leverage the framework of receding horizon control and propose an interaction-aware motion prediction model that can make accurate and reactive predictions of surrounding traffic agents to support the decision-making process.

\subsection{Motion Prediction}
Motion prediction is a thread of research that aims to predict long-term future motion trajectories of traffic participants based on their historical dynamic states and optionally the map information. With the introduction of Transformers or GNNs, recent motion prediction networks have gained the ability to effectively handle a heterogeneous mix of traffic entities, e.g., road polylines, traffic light state, and a dynamic set of agents, and achieved unprecedented prediction accuracy \cite{huang2022multi, ngiam2021scene, salzmann2020trajectron++, varadarajan2022multipath++}. However, most of the existing motion prediction models only focus on improving the prediction accuracy (i.e., position error), ignoring the applicability to the downstream planning task. One prominent issue is that the model is not aware of the AV's future plans and the prediction results are not reactive to the AV's different decisions, forcing the AV to act passively. Some works have recognized this issue and made some efforts to mitigate it \cite{espinoza2022deep, song2020pip, tolstaya2021identifying, huang2022conditional, tang2022interventional}. PiP \cite{song2020pip} proposes a planning-informed trajectory prediction network that conditions the prediction process on the candidate trajectories of the AV, and \cite{tolstaya2021identifying} formulates such a framework as conditional behavior prediction. More recently, \cite{tang2022interventional} argues that the AV's future plan should be treated as an intervention rather than an observation in the prediction model. \cite{espinoza2022deep} proposes a multi-agent policy network as the prediction model that can react to the AV's action. Nevertheless, they are trained with offline driving datasets and still focus on the prediction part while the decision-making performance and interactive behaviors are less investigated. Our work strives to train the interaction-aware prediction model online through interacting with the other agents and thoroughly evaluate the decision-making performance of the interactive prediction model.

\section{Methodology}
\subsection{Decision-making Framework} 
\begin{figure}[htp]
    \centering
    \includegraphics[width=\linewidth]{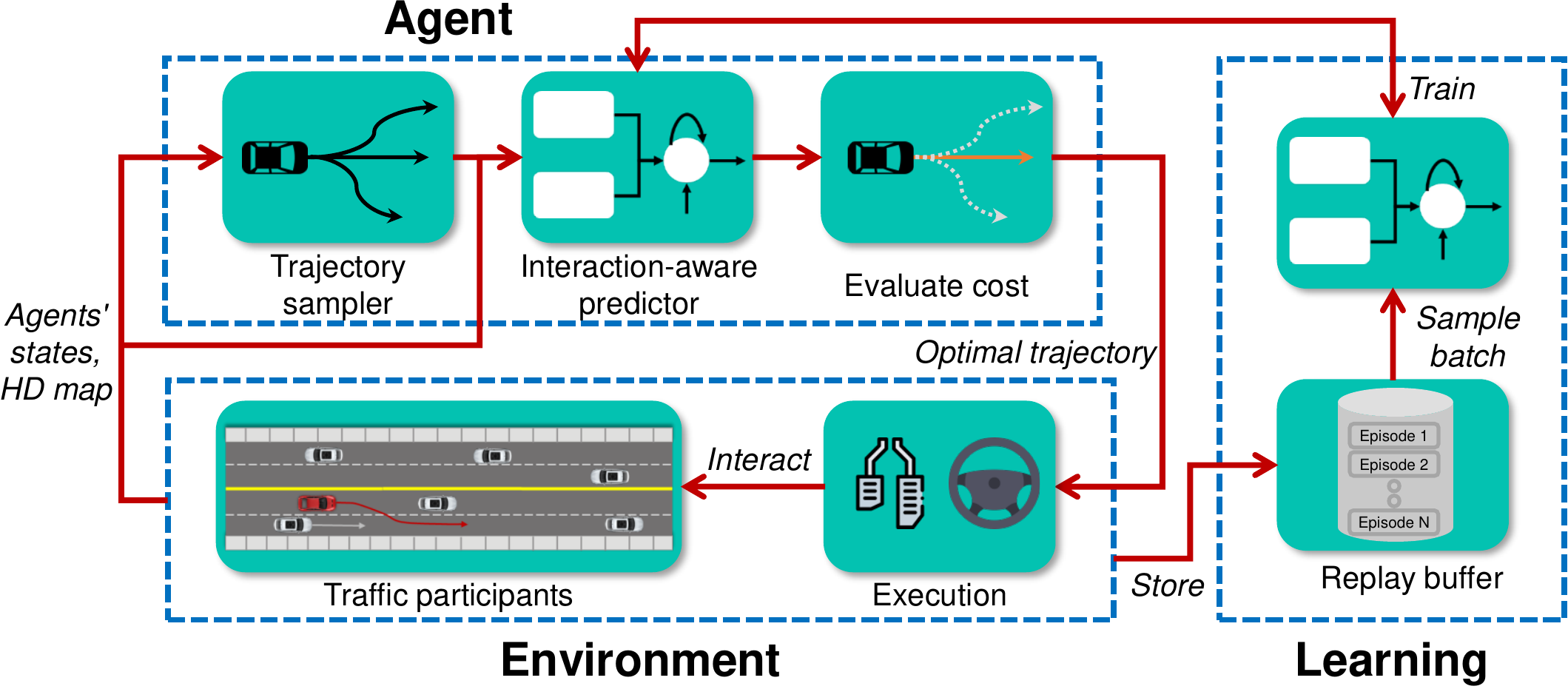}
    \caption{An overview of the decision-making and online learning framework. The agent runs the proposed decision-making framework that generates candidate trajectories, predicts other agents' reactions, selects the optimal one, and executes it until the next cycle, to interact with other traffic participants in the environment. The episodic data is stored in a replay buffer and batch data is sampled from it to train the motion predictor.}
    \label{fig:fig.1}
\end{figure}

The goal of the decision-making framework is to generate safe, efficient, and comfortable trajectories for the AV to follow, considering the interactions with surrounding agents. We leverage the receding horizon control scheme, where a trajectory is planned within a finite horizon and executed until a new trajectory is replanned. Let $\mathbf{x}^{e}_t$ denote the state of the AV and $\mathbf{x}^{\neg e}_t$ the states of the AV's surrounding agents at timestep $t$, and $f$ is the prediction model that forecasts the future state sequences of other agents $\mathbf{\hat x}^{\neg e}_{1:T}$, the decision-making framework selects the optimal trajectory $\tau^{*}$ (i.e., a sequence of the AV's future states $\mathbf{x}^e_{1:T}$) that minimizes the expected cost:
\begin{equation}
\begin{split}
\tau^{*} &= \underset{\tau \in \mathcal{T}} {\mathrm{argmin}} \ c(\mathbf{x}^e_{1:T}, \mathbf{\hat x}^{\neg e}_{1:T}), \\
\mathbf{\hat x}_{1:T} &= f(\mathbf{x}^{\neg e}_{-T_h:0}, \mathcal{I}, \mathbf{x}^e_{1:T}),
\end{split}
\end{equation}
where $T$ is the planning horizon, $T_h$ is the historical horizon, and $\mathcal{I}$ is the map information. $\mathcal{T}$ is a set of feasible trajectories generated based on the initial AV state $\mathbf{x}^{e}_0$, and $c$ denotes the cost function. The prediction result can be a distribution of possible future states per agent or joint trajectories for multiple agents. Still, for simplicity in this paper, the result would be a single trajectory per agent. The decision-making framework and learning process are illustrated in Fig. \ref{fig:fig.1} and the key components are detailed as follows.

\begin{figure}[htp]
    \centering
    \includegraphics[width=\linewidth]{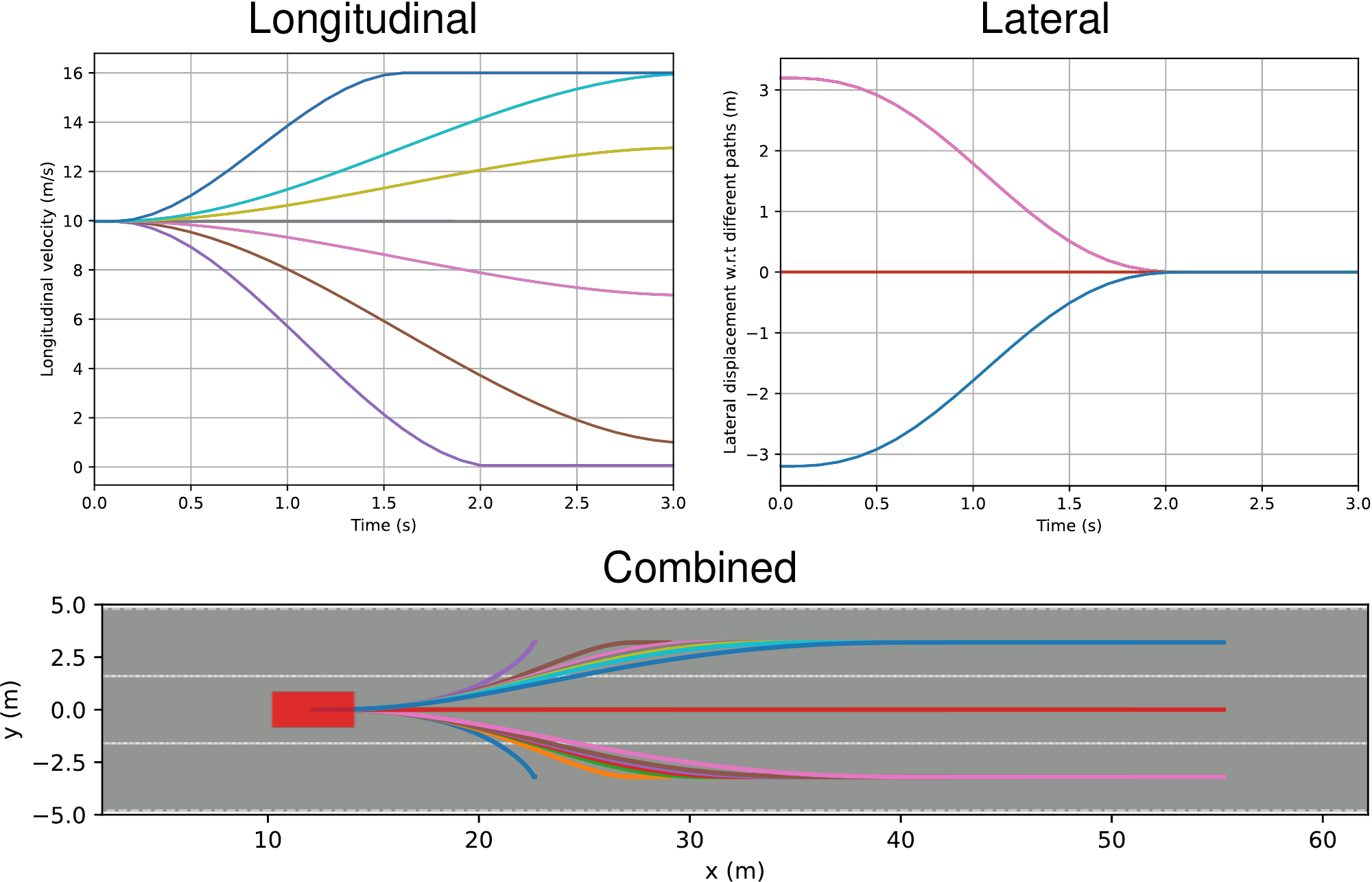}
    \caption{The trajectory generation process. Speed profiles in the longitudinal direction and displacement profiles in the lateral direction are generated and combined to obtain the candidate trajectories.}
    \label{fig:fig.2}
\end{figure}

\textbf{Trajectory generation}. 
The framework first generates a set of candidate trajectories based on the AV's current state. We decouple the trajectory into longitudinal and lateral directions in the Frenet path \cite{werling2010optimal}. In the longitudinal direction, we utilize cubic polynomials to generate different speed profiles from the initial speed to different target speeds, and the speed remains unchanged after reaching the target speed. The longitudinal coordinates can be obtained by integrating the speed profile over time. In the lateral direction, we employ a quintic polynomial to produce the trajectory changing to the target lane within a target time and staying in the target lane afterward. Then, the longitudinal and lateral trajectories are combined and translated back to Cartesian space. Fig. \ref{fig:fig.2} shows an example of the trajectory generation process in a typical driveway with three candidate paths.

\textbf{Evaluation}. 
The cost function to evaluate the candidate trajectories is designed as a linear combination of different carefully crafted features that characterize driving behaviors:
\begin{equation}
c(\mathbf{x}^e_{1:T}, \mathbf{\hat x}^{\neg e}_{1:T}) = \sum_i \omega_i \mathrm{f}_i(\mathbf{x}^e_{1:T}, \mathbf{\hat x}^{\neg e}_{1:T}),
\end{equation}
where $\omega_i$ is the parameter, $\mathrm{f}_i$ is the feature of the AV's plan that may involve predicted trajectories of other agents.

Regarding the features, we consider the factors of safety, ride comfort, travel efficiency, and traffic rules. The safety factor includes three features, which are collision $\mathrm{f}_{col}$, distance to other agents $\mathrm{f}_{d2a}$, and time to collision (TTC) $\mathrm{f}_{ttc}$. The calculation of these features requires the predicted trajectories of surrounding agents. The planned trajectories that overlap with the predicted trajectories of other agents in the Frenet frame, at a close distance to them, or with dangerous TTC values are heavily penalized. We promote high travel speed but not going above the speed limit, and thus $\mathrm{f}_{spd}$ is the absolute difference between the vehicle speed and speed limit. We also utilize the distance to the goal lane $\mathrm{f}_{tgt}$ to encourage the AV to stay close to the target lane. For ride comfort, longitudinal jerk $\mathrm{f}_{jerk}$ and lateral acceleration $\mathrm{f}_{acc}$ are selected. In summary, there are seven features in the cost function, and the cost parameters are carefully tuned to deliver expected behaviors.

\textbf{Execution}. 
We employ a low-level motion controller to execute the first few steps of the trajectory and then replan. In the training process, to better reflect the influence of the plan on other agents, we execute longer steps of the planned trajectory. In testing, we execute fewer steps to ensure the safety and adaptability of the framework.

\subsection{Prediction Model}
\begin{figure*}[htp]
    \centering
    \includegraphics[width=0.9\linewidth]{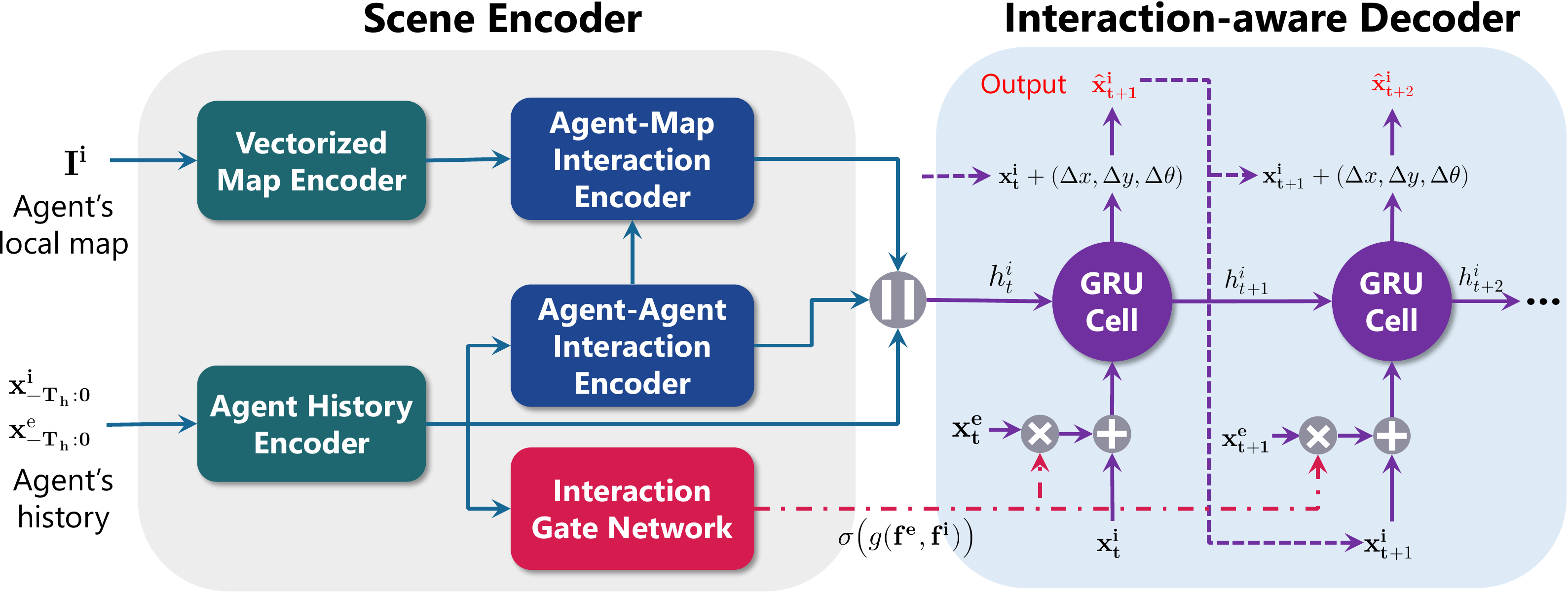}
    \caption{An overview of our proposed interaction-aware motion prediction model. The agent's history and local map are encoded with the scene encoder to obtain an initial hidden state of the agent; a GRU-based decoder is used to predict the agent's future state incorporating the AV's planned trajectory.}
    \label{fig:fig.3}
\end{figure*}

The prediction model takes as input the historical trajectories and local map polylines of surrounding agents and jointly outputs their future trajectories. In addition, we incorporate the AV's planned trajectory into the model, so that the model can predict the reactions of other agents to the AV's plan. Fig. \ref{fig:fig.3} depicts the structure of the proposed motion prediction model and the two main parts, i.e., scene encoder and interaction-aware decoder, are elaborated as follows. 

\textbf{Scene Encoder}. 
The scene encoder is based on Transformer networks, following the structure of our previous works \cite{huang2022multi, huang2022differentiable}. For each surrounding agent $i$, the input data consists of its historical states $\mathbf{x}^i_{-T_h:0}$ and local map polylines $\mathcal{I}^i$ (i.e., a list of waypoints of nearby routes). We also include the AV's historical states $\mathbf{x}^e_{-T_h:0}$ to yield a complete interaction graph. The missing timesteps in the historical trajectory and missing points in map polylines are padded as zeros to keep the tensor shapes fixed. We select a fixed number of agents closest to the AV and pad with zeros if there are too few. The raw map polylines are encoded using a shared vectorized map encoder, which is a multi-layer perceptron (MLP). The historical states of agents are encoded using a shared agent history encoder, which is a self-attention Transformer layer, to extract the temporal relation on the agent's historical trajectory. We use the feature at the last timestep from the encoder as a latent representation of the agent and employ hierarchical Transformer modules to encode the agent-agent and agent-map relations. Specifically, all agents' features are gathered as nodes in the interaction graph (all agents are connected), and a self-attention Transformer module is used to process the graph to capture the relationship between the interacting agents. Then, we utilize a cross-attention Transformer layer with the query as the interaction feature of the agent and key and value as the feature vectors of the map polylines, to capture the relationship between the agent and map. Finally, for each surrounding agent, its historical feature, interaction feature, and map attention feature are concatenated and passed through the interaction-aware decoder to generate the predicted future trajectory.

\textbf{Interaction-aware Decoder}.
We employ the gated recurrent unit (GRU) cell to decode a sequence of future states of the agent in an autoregressive manner. Importantly, we incorporate the AV's planned state $\mathbf{x}^e_t$ into the cell to model the influence of the AV's action on the agent's future state and make interaction-aware predictions. Specifically, at each future timestep, the GRU cell receives the hidden state from the last timestep and joint input of the last predicted state and the AV's future state at that timestep. The updated hidden state is used to decode the change of the agent state (including $x$ and $y$ coordinates and heading angle $\theta$), which is added up to the last state to yield the predicted state at the current timestep. Moreover, since the AV may not impose influence on some agents that are not in the conflict zone, we add an interaction gate network $g$ to model the pair-wise relation between the AV and a target agent. The gating network, which is an MLP, takes as input the concatenated features of the AV and target agent from the agent history encoder and outputs a score through a sigmoid function to indicate the interaction intensity between the AV $e$ and agent $i$, $\sigma(g(\mathbf{f}^e, \mathbf{f}^i))$. The intensity will be applied to the input of the AV's future states channeled to the GRU cell. 

\subsection{Online Learning} 
We design an online learning procedure to train the framework, which is summarized in Algorithm \ref{alg1}. During the online interaction, the agent complies with the decision-making framework to generate candidate trajectories, evaluate them, and execute the optimal trajectory $N_e$ steps. To enable the agent to explore the elicit other agents' reactions, we set up a parameter $\epsilon \in [0, 1]$ denoting the probability the agent would ignore the safety cost and greedily takes the actions to reach the target. The parameter $\epsilon$ would gradually decrease from 1 (totally ignoring safety) during the training process. We keep an episodic buffer to store the agents' trajectories under the global coordinate system and the episodic trajectories are dumped into a replay buffer at the end of an episode.

We train the motion predictor at the end of every episode with $N_t$ gradient steps. At each gradient step, we sample a batch of data from the replay buffer from different episodes and timesteps. We can obtain the historical trajectories of the agents and their future trajectories at each timestep from the episode memory, as well as their local map polylines from an offline database. We use the AV's recorded future trajectory as the planned AV trajectory input to the motion predictor. The position and heading attributes of the input data are normalized according to the AV's state at that timestep. The data point is added to a mini-batch with a size of $N_b$ and the loss function for a mini-batch is: 

\begin{equation}
\label{loss}
\mathcal{L} = \frac{1}{N_b N_a} \sum_{i=0}^{N_b} \sum_{j=1}^{N_a} \mathcal{L}_{SL_1} \left(\mathbf{\hat x}^{j}_{i}, \mathbf{x}_{i}^{j} \right),
\end{equation}
where $N_a$ is the number of surrounding agents, $\mathcal{L}_{SL_1}$ is the smooth L1 loss, $\mathbf{\hat x}^{j}_{i}$ is the predicted state of agent $j$, and $\mathbf{x}^{j}_{i}$ is the ground truth state. 

\RestyleAlgo{ruled}
\SetKwComment{Comment}{/* }{ */}
\SetKwInOut{Input}{Input}

\begin{algorithm}[hbt!]
\caption{Online learning of the decision-making framework with interaction-aware motion prediction}
\label{alg1}
\Input{cost function parameter $\omega$, episodic buffer $\mathcal{D}_e$, empty replay buffer $\mathcal{D}$, motion prediction network $\theta$}
\For{Episode $\gets 1$ \KwTo $N_e$ }{
  Reset environment state\;
  Empty the episodic buffer $\mathcal{D}_e$\;
  \While{environment state is not terminal}{
    Generate a set of trajectories $\mathcal{T}$\;
    With probability $\epsilon$ set the safety cost to $0$\;
    \ForEach{$\tau \in \mathcal{T}$}{
    Query the model with agent history, map, and AV plan $\tau$ to get prediction results\;
    Calculate the cost of plan $\tau$\; 
    }
    Execute $N_e$ steps of the optimal trajectory in the environment and store agents' states into episodic buffer $\mathcal{D}_e$ every step\;
  }
  Dump episodic buffer $\mathcal{D}_e$ into replay buffer $\mathcal{D}$\;
  \For{Gradient\_step $\gets 1$ \KwTo $N_t$}{
     Initialize a mini-batch $\mathcal{B}$\;
      \For{$i \gets 1$ \KwTo $N_b$}{
        Sample an episode from replay buffer $\mathcal{D}$ and a timestep from the episode\;
        Process the data and add to mini-batch $\mathcal{B}$\;
      }
      Calculate loss according to Eq. (\ref{loss})\;
      Update the model parameter $\theta$\;
    }
}
\end{algorithm}

\section{Experiments}
\subsection{Experimental Setup}
\subsubsection{Driving Scenarios}
\begin{figure}[htp]
    \centering
    \includegraphics[width=\linewidth]{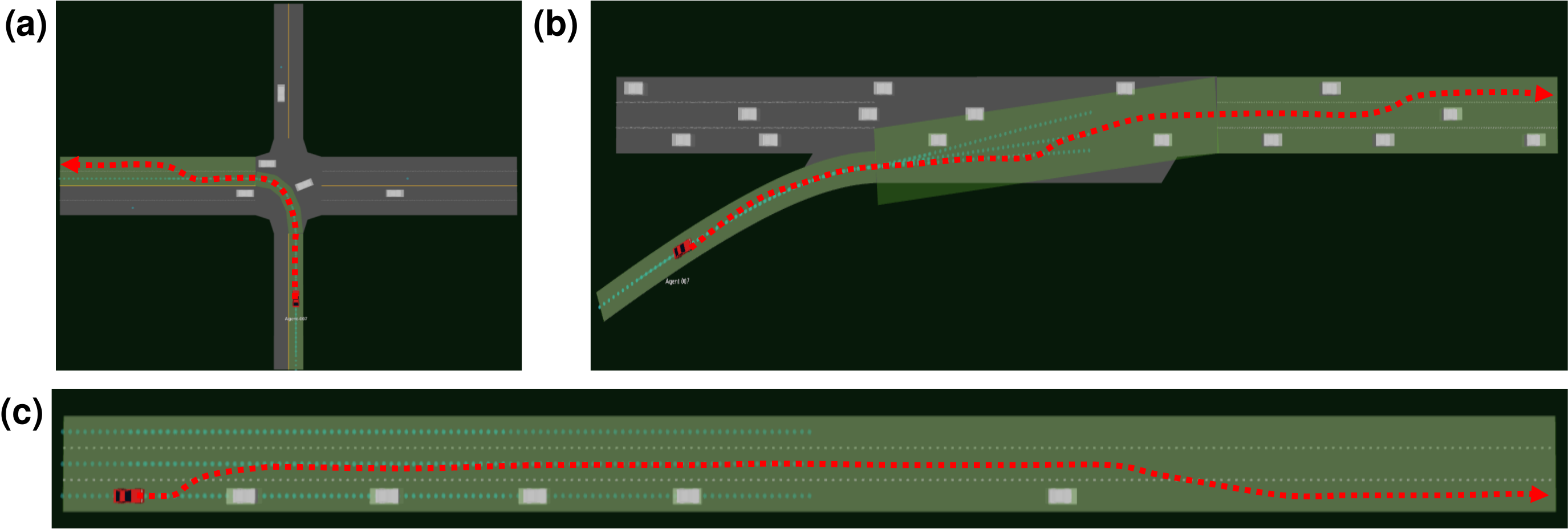}
    \caption{Three designed interactive driving scenarios in the SMARTS environment: (a) intersection; (b) merge; (c) overtake.}
    \label{fig:fig.4}
    \vspace{-0.3cm}
\end{figure}

We design three challenging and highly interactive driving scenarios in the SMARTS simulator \cite{zhou2020smarts}, which are displayed in Fig. \ref{fig:fig.4}. In the unsignalized \textbf{intersection} scenario (Fig. \ref{fig:fig.4}(a)), the AV needs to handle the traffic flows from different directions and coordinate with other agents to safely and efficiently pass the intersection. In the \textbf{merge} scenario (Fig. \ref{fig:fig.4}(b)), the AV is tasked to change to the leftmost lane starting from the ramp and safely navigate through the dense traffic on the main road. In the \textbf{overtake} scenario (Fig. \ref{fig:fig.4}(c)), the AV is supposed to overtake the slow-moving vehicles in the current lane and change back to the original lane at the end of the segment. We also generate a large variety of traffic flows in each scenario to ensure the diversity of the data. All the designed scenarios require accurate and interactive predictions of surrounding agents to enable the AV to make safe and human-like decisions. 

\subsubsection{Baseline methods}
To validate the performance of our proposed framework, we set up some baseline methods. \textbf{RL methods}: PPO \cite{schulman2017proximal}, SAC \cite{haarnoja2018soft}, and TD3 \cite{fujimoto2018addressing} are employed to train the agent. The action is the target speed and lane, and the feature extractor of policy and value networks is a simplified version of the prediction network (only the multi-head attention modules are kept). The reward function is the negative value of the step-wise cost function previously designed. \textbf{Ablated methods}: three sub-models with some key components of the framework removed are used to investigate their functions. First, the interaction-aware part in the prediction model is ablated, meaning the future plan of the AV will not be input to the GRU cell. Second, we remove exploration in the online learning process. Third, we replace the prediction model with a non-learnable constant velocity and turn rate (CVTR) model. 


\subsection{Implementation Details}
The prediction and planning horizon in our framework is 3 seconds ($T=30$) with a time interval of 0.1 seconds, and the historical observation horizon of an agent is 1 second ($T_h=10$). We search for 5 neighboring agents closest to the AV and predict their future trajectories. For each agent, we find three nearby routes as the vectorized local map, each with 50 waypoints and a space interval of 1 meter. The zero-padded historical timesteps, map waypoints, and agents are masked from attention calculations in the network. The trajectory with the smallest cost from the candidate trajectory set is selected and executed for 15 steps in the training process and 5 steps in testing. 

We use the Adam optimizer in PyTorch to train the prediction model, and the learning rate starts with 2e-4 and decays by a factor of 0.8 after every 5000 gradient steps. The number of training episodes is 1000, the number of gradient steps after every episode is 50, the batch size is 32, and the exploration rate $\epsilon$ decays from 1 to 0.05 after 500 episodes. All the learnable methods are trained five times with different random seeds and our framework takes about 8 hours with AMD Ryzen 3900X CPU and NVIDIA RTX 3080 GPU for one training process. 

\subsection{Results}

\subsubsection{Training}
\begin{figure*}[htp]
    \centering
    \includegraphics[width=0.98\linewidth]{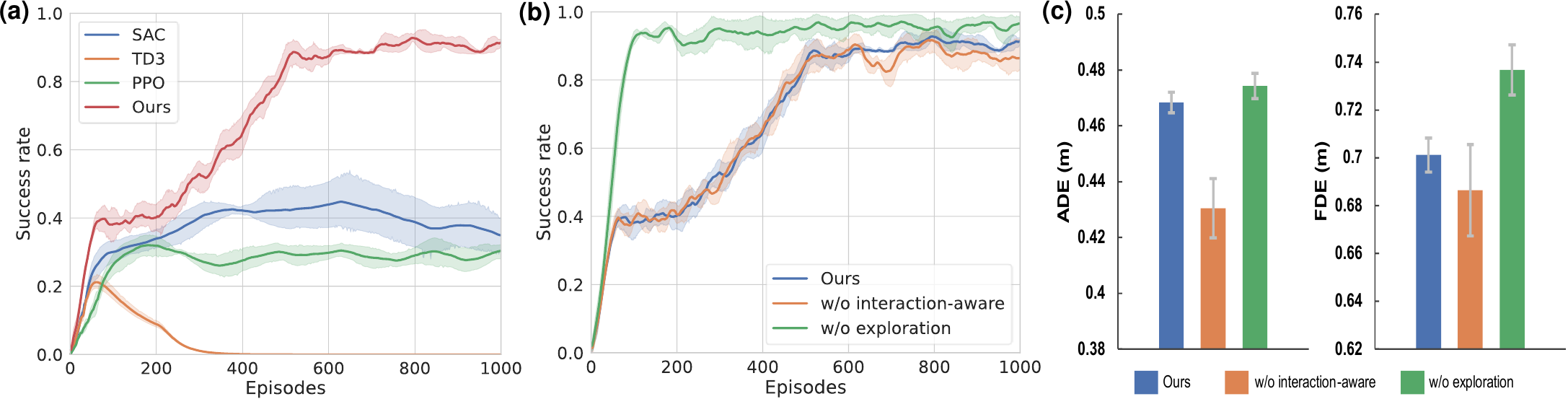}
    \caption{The training results of the proposed decision-making framework and baseline methods: (a) comparison with RL methods; (b) comparison with ablated methods; (c) comparison of motion prediction accuracy.}
    \label{fig:fig.5}
\end{figure*}

We train the decision-making framework in the three interactive driving scenarios simultaneously. Fig. \ref{fig:fig.5} shows the training process of the proposed framework together with RL-based methods and ablated methods, as well as the prediction performance of the model. The metric for decision-making performance is the average success rate, and the metrics for prediction performance are average displacement error (ADE) and final displacement error (FDE). The results in Fig. \ref{fig:fig.5}(a) suggest that our method has a superior performance in data efficiency and success rate over the RL methods. While our method can steadily reach a success rate of over 90\% at the end of the training, the RL methods suffer from poor performance, especially the TD3 algorithm (the agent not moving anymore). Fig. \ref{fig:fig.5}(b) demonstrates that the proposed interaction-aware prediction method can bring a higher success rate than the non-interaction-aware method and removing the exploration process can quickly reach a high and stable success rate. However, the results in Fig. \ref{fig:fig.5}(c) indicate that the prediction accuracy of the interaction-aware method is not improved. One possible explanation is that adding the AV's future trajectory to the decoder can cause the prediction mistakenly dependent on the AV's future plan or result in causal confusion, but this problem can be mitigated by collecting more diverse training data through the exploration process.

\begin{figure*}[htp]
    \centering
    \includegraphics[width=0.9\linewidth]{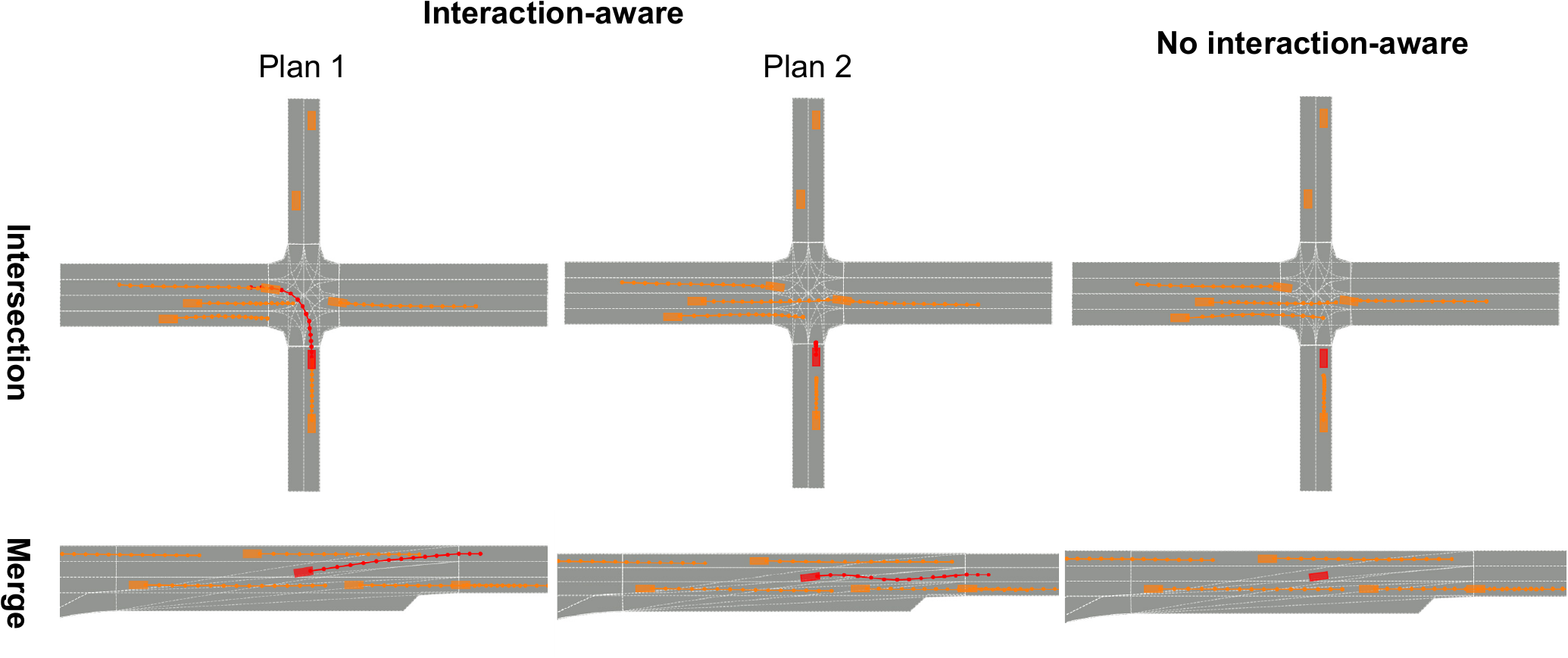}
    \caption{The prediction results of the interaction-aware model and no interaction-aware model. The red box and line are the AV and its planned trajectory; the orange boxes and lines are the surrounding agents and their predicted trajectories.}
    \label{fig:fig.6}
\end{figure*}

\subsubsection{Testing}
We use 50 different traffic flows for each scenario to test the performance of the decision-making framework in comparison with the baseline methods. Table \ref{tab1} reports the testing results of our proposed method and several RL methods, which indicate that our method significantly outperforms the RL methods in terms of both success rate and time efficiency. Notably, no RL method can finish the merge task because it requires complex interaction with other agents and a clear awareness of the goal. In the overtake task, although the PPO method has a high success rate, the agent only learns to follow the front vehicle instead of overtaking, thus having a longer average travel time. The results also manifest the limitations of RL approaches in driving tasks. Table \ref{tab2} compares our method with ablated methods to investigate the effects of these components on decision-making performance. The results reveal that learning-based prediction models outperform the kinematic-based model, especially in scenarios with complex road structures, such as intersections. Compared to the non-interaction-aware model, our proposed method shows a better success rate and lower collision rate. In some cases, the non-interaction-aware method fails to change to the target lane, which is occupied by another agent, but our method can make reactive planning and handle such situations. Moreover, the exploration process could bring better testing performance, as a result of collecting more diverse data on other agents' reactions to the AV to better train the model. 

\begin{table}[htp]
\caption{Testing results in comparison with RL methods}
\resizebox{\linewidth}{!}{%
\begin{tabular}{c|cccccc|c}
\toprule
\multirow{2}{*}{Method} & \multicolumn{2}{c}{Intersection} & \multicolumn{2}{c}{Merge} & \multicolumn{2}{c|}{Overtake} & \multirow{2}{*}{\begin{tabular}[c]{@{}c@{}}Overall\\ Success\end{tabular}} \\
                        & Success          & Time          & Success       & Time      & Success        & Time        &                    \\ \midrule
TD3                     & 6\%              & 13.5s         &  0\%          & --        &  12\%          &  22.5s      & 6.0\%                         \\
SAC                   & 54\%             & 10.1s         &  0\%          & --        &  40\%          &  17.8s      & 31.3\%                    \\
PPO                    & 76\%             & 15.1s         &  0\%          & --        &  98\%          &  27.2s      & 58.0\%                    \\
Ours                    &\textbf{98}\%     & 11.9s         &\textbf{100}\%  & 14.3s     &\textbf{100}\%  &  16.3s      &  \textbf{99.3}\%         \\ \bottomrule
\end{tabular}
}
\vspace{-0.3cm}
\label{tab1}
\end{table}

\begin{table}[htp]
\caption{Testing results in comparison with ablated methods}
\resizebox{\linewidth}{!}{%
\begin{tabular}{c|cccccc|c}
\toprule
\multirow{2}{*}{Method} & \multicolumn{2}{c}{Intersection} & \multicolumn{2}{c}{Merge} & \multicolumn{2}{c|}{Overtake} &  \multirow{2}{*}{\begin{tabular}[c]{@{}c@{}}Overall\\ Success\end{tabular}} \\
                        & Success          & Collision     & Success        & Collision      & Success        & Collision  &                          \\ \midrule
w/ CVTR                 &  78\%            & 22\%          & 90\%           &  8\%           &  94\%          & 6\%        & 87.3\%                 \\
w/o exploration         &  94\%            & 4\%           & 98\%           &  2\%           &  98\%          & 2\%        & 96.7\%                        \\
w/o interaction         &  94\%            & 4\%           & 94\%           &  4\%           & \textbf{100}\% &  0\%       & 96.0\%          \\
Ours                    &  \textbf{98}\%   & 2\%           &\textbf{100}\%  &  0\%           & \textbf{100}\% &  0\%       & \textbf{99.3}\%          \\  \bottomrule
\end{tabular}
}
\label{tab2}
\end{table}

\subsubsection{Interactive Prediction}
Fig. \ref{fig:fig.6} shows some examples of our framework's ability to make interactive predictions. In the intersection scenario, given two different plans of the AV (pass and yield), the model delivers two distinct prediction results of surrounding agents reacting to the AV's plan. Only the agents that have a direct conflict with the AV are influenced and other agents' predicted trajectories are unchanged. In the merge scenario, our model can also predict the agent's reaction on the target lane to the ego agent's merge attempt, while the non-interaction-aware model only outputs a fixed result, which could degrade the decision-making performance in highly interactive scenarios.

\section{CONCLUSIONS}
In this paper, we propose a decision-making framework based on an interaction-aware motion prediction model. The prediction model utilizes Transformers to encode the driving scene and incorporates the AV's planned trajectory in decoding to reflect the influence of the AV on other agents. We design an online learning framework to train the model and validate it in simulated driving scenarios. The results demonstrate that our method significantly outperforms the baseline RL methods, and using an interaction-aware model achieves a higher success rate than a non-interaction-aware model or without the online exploration process. 

\bibliographystyle{IEEEtran}
\bibliography{IEEEexample}
\end{document}